\documentclass{article}

     \usepackage[preprint]{neurips_2020}

\usepackage[utf8]{inputenc} 
\usepackage[T1]{fontenc}    
\usepackage{hyperref}       
\usepackage{url}            
\usepackage{booktabs}       
\usepackage{amsfonts}       
\usepackage{nicefrac}       
\usepackage{microtype}      
\usepackage{graphicx}

\title{Convolutional-LSTM for Multi-Image to Single Output Medical Prediction}

\author{%
  Luis Fernando Leal \\
  \texttt{wichofer89@gmail.com} \\
   \And
   Marvin Castillo \\
   \texttt{mcastilloy2k@gmail.com} \\
   \And
   Fernando Juarez \\
   \texttt{frenandister@gmail.com} \\
   \And
   Erick Ramirez \\
   \texttt{erickramireztebalan@gmail.com} \\
   \And
   Mildred Aspuac \\
   \texttt{mildred.aspuac@gmail.com} \\
   \And
   Diana Letona \\
   \texttt{Dmletona@gmail.com} \\
}

\begin{document}

\maketitle

\begin{abstract}  Medical head CT-scan imaging has been successfully combined with deep learning for medical diagnostics of head diseases and lesions[1]. State of the art classification models and algorithms for this task usually are based on 3d convolution layers for volumetric data on a supervised learning setting (1 input volume, 1 prediction per patient) or 2d convolution layers in a supervised setting (1 input image, 1 prediction per image). However a very common scenario in developing countries is to have the volume metadata lost due multiple reasons for example  formatting conversion in images (for example .dicom  to jpg), in this scenario the doctor analyses the collection of images and then emits a single diagnostic for the patient (with possibly an unfixed and variable number of images per patient) , this prevents it from being possible to use state of the art 3d models, but also is not possible to convert it to a supervised problem in a  (1 image,1 diagnostic) setting because different angles or positions of the images for a single patient may not contain the disease or lesion. In this study we propose a  solution for this scenario by combining 2d convolutional[2] models with  sequence models which generate a prediction only after all 
images have been processed by the model for a given patient  \(i\), this creates a multi-image to single-diagnostic setting \(y^i=f(x_1,x_2,..,x_n)\) where \(n\) may be different between patients. The experimental results demonstrate that it is possible to get a multi-image to single diagnostic model which mimics human doctor diagnostic process: evaluate the collection of patient images and then use important information in memory to decide a single diagnostic for the patient. 
\end{abstract}

\section{Introduction}

\subsection{Motivation and Justification}
Head CT-scans are one of the most common medical studies for medical diagnostics, for some head diseases and lesions a fast diagnostic from CT-scans can be crucial for potentially saving patient lives. In developing countries where a lack of specialists create bottlenecks it can be very useful to have a diagnostic system to aid  and help  doctors and/or serve as a second opinion for them. Advances in deep learning have enabled training advanced computer vision models for this task[1], however state of the art models usually work on one of the following settings:

\begin{itemize}

\item \textbf{Setting 1}
 Supervised learning setting having one diagnostic per 2d image: if there are many images for a single patient, every image has a diagnostic , then the doctor summarizes all findings for a final diagnostic. This requires 1 diagnostic label for every input image for doing supervised learning meaning training examples are in the form: 
  \begin{center}
 $(x,y)$ where x= one patient image, y = diagnostic for that image.
\end{center}
Deep learning models exist[1] for this setting where the model gets a prediction per image and then all are summarized for a final prediction per patient.

\item \textbf{Setting 2}
Supervised learning setting having one diagnostic per 3d volume: using special imaging formats (for example .dicom) that contain volumetric metadata a 3d volume can be retrieved from multiple 2d images (an associated volumetric metadata), the doctor then emits a diagnostic for the 3d volume (not every 2d image). This requires one diagnostic label for every input 3d volume (not the underlying 2d images that make up the volume)
meaning training examples are in the form: 
  \begin{center}
 $(x,y)$ where x= one patient volume(from 2d slices), y = diagnostic for that patient.
\end{center}
\end{itemize}
A similar approach to this is used by Casamitjana et tal[4] for image segmentation instead of classification.

However in developing countries it is common that for many reasons (including scarce computer disk and storage) 2d images that make up 3d volumes are converted to a different format (for instance from .dicom to .jpg) losing 3d volumetric metadata in the process and thus rolling to a collection of 2d images per patient. This prevents it from being possible to apply any of the methods previously outlined:

\begin{itemize}

\item \textbf{Setting 1}
Every patient \(i\) has multiple 2d images, but there is no a diagnostic per image, just a general diagnostic for the patient . The patient diagnostic cannot be replicated to all the \(n\) images because some slices will not contain the disease or lesion.

\item \textbf{Setting 2} 
As stated during format conversion volumetric metadata is lost making impossible to reconstruct the 3d volume from the 2d slices , thus a 3d model cannot be applied.
\end{itemize}

Given the positive impact a computer vision model can have for a task as critical as early head disease/lesion detection, specially in countries where the number of medical specialists is very low,we believe an approach for training a computer vision model that solves the issues stated ( commonly unknown by big research labs) can be valuable. 
\subsection{Study Hypothesis}
Inspired by the medical doctor approach: Visually analyze the set of 2d images for a given patient \(i\) and use important information and facts stored in memory for emitting a diagnostic for the patient ,we formulate our work hypothesis: \emph{If a vision model is combined with a memory sequence model , then it will be able to store relevant visual features from a sequence of images in order to emit a single diagnostic for the collection of images}.

More formally the hypothesis states that deep learning can provide the tools to approximate a function that receives a sequence $S$ of vectors $s$(every $s$ being a non-linear vector valued function of an 2d image $s(x_i))$  

\begin{center}
  $S=(s(x_1),..,s(x_{T{x}})))$
\end{center}

 and outputs a prediction $y$ 
 \begin{center}
   $y = f(S)$
\end{center}

 where f is implemented based on an auto-regressive non-linear function 
  \begin{center}
   $h_t= g(s(x_t),h_{t-1})$
\end{center}

and $h_t$ is a hidden state at time $t$ that depends on the input vector $s(x_t)$ and the hidden state at time $t-1$. The final output $y$ is a function(also non-linear) of g obtained at the last time-step of the sequence $x_{T{x}}$ :
  \begin{center}
   $y = f(g(s(x_{T{x}})))$
\end{center}

For this project a convolutional[2] neural network is used to approximate $s(x)$    and  a LSTM[3] neural network is used to approximate the function $g$

Translated to deep learning terminology the hypothesis states that a convolutional model for 2d images combined to a sequence model can be designed in a many-to-one setting of multiple image input to single diagnostic output using the sequence model to store relevant features for the complete sequence of images, these features then are used to get a single diagnostic for a given patient taken  from a fully connected layer connected to the LSTM only for the last image of the sequence. 
\subsection{Objective}
The main goal of this study is to design, develop and test deep learning models combining vision capabilities and memory/sequence models for the multi-image to single diagnostic medical task, given this is an independent research project using a small data-set(kindly provided by a small medical lab from from Guatemala) experiments are limited to personal hardware resources including a single computer and a single GPU, thus small feasible candidate models are selected instead of state of the art models non feasible for single GPU, thus the aim of this study is to validate the hypothesis and provide a proposed solution for the issues outlined, not to get state of the art accuracy in the medical diagnostic setting.

We hypothesize better experimental results and better models can be obtained from the use of bigger state of the art  models and bigger data-set.

\section{Methods}
For this study we performed a deductive experimental approach that consists of designing the deep learning models(including selection of architectures from the literature) and training algorithm, running multiple experiments varying the architecture and/or hyper-parameters and capturing the training and validation performance metrics, we report just the more relevant experiments and results.

Independently of any specific architecture, design or hyper-parameters there some common details in our experimental setup described here:

\paragraph{Data-set}
The data-set used was provided(and authorized us to use it) by a local medical clinic which specializes in radiology, the data-set contains a collection of 2d images as well as the clinical record for every patient, label extraction was performed manually by a medical doctor(part of the project team) by reading the clinical record of every patient and recording the diagnostic on a file with 1 diagnostic per patient.

The complete data-set contains data belonging to 1142 patients(anonymized), every patient having a set of 2d images(variable number of images across patients) for a total of 16504 images(having an average of 23 images per patient), the final data-set is smaller due to filtering(which is described in the following section) for a final data-set size of 715 patients.

The data-set(after manual labeling) provides one diagnostic per patient where every diagnostic it's made up of  an indicator variable for 5 possible diseases, table 1 summarizes the prevalence for every disease.

\begin{table}
  \caption{Disease prevalence}
  \label{disease-prevalence}
  \centering
  \begin{tabular}{lll}
    \cmidrule(r){1-2}
    Disease     & Prevalence      \\
    \midrule
    Brain Hemorrhage & 0.2830    \\
    Brain Ischemia     & 0.1120      \\
    Brain Fracture     & 0.0715       \\
    Brain Mass     & 0.0187       \\
    Brain Edema     & 0.1540        \\
    \bottomrule
  \end{tabular}
\end{table}

For the final results (reported in the results section and related experiments described in this section) the data-set was split(by patient) in train and validation/model selection using 643 patients for training and 72 patients for validation and model selection, due to the small data-set size and not having high generalization performance as a goal of this particular study we did not use a test set for measuring final generalization error and we plan to do that on follow up work.

\paragraph{Pre-Processing}

Performed pre-processing tasks on the data-set  are listed next:

\begin{itemize}

\item Patients with no diagnostic were removed.
\item Patients having more than 50 images were removed due to memory exhaustion on the GPU.
\item Images were converted from  gray-scale to RGB by replicating the depth channel 3 times to be able to use pre-trained models for transfer learning.

\end{itemize}
\paragraph{Data Augmentation}

Due to the small size of the data-set in order to prevent over-fitting we performed artificial data synthesis through data augmentation by applying small random transformations on the original images including: 
\begin{itemize}
\item Random Rotation: random rotation of 10 degrees max
\item Random Resized Crop: zoom-in up to 60%
\end{itemize}
An example can be seen in \textbf{figure 1}

\begin{figure}
    \centering
    \includegraphics[height=5cm]{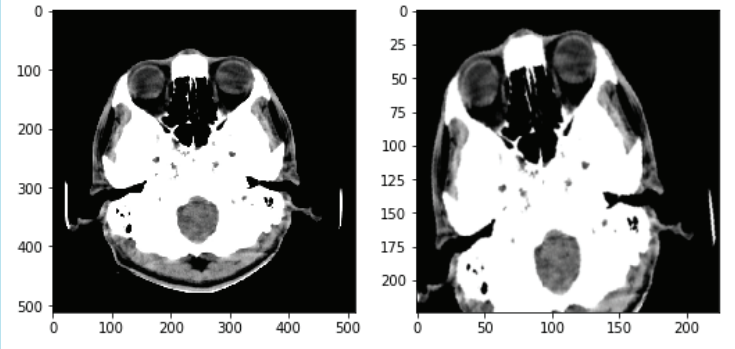}
    \caption{Original image(left) vs random augmented image(right)}
\end{figure}

\paragraph{Stochastic gradient descent based optimization(but accumulating gradients)}
For this project we wanted the model to be able to have a variable-length input sequence capability(instead of having a fixed size window having some patients truncating some images), this derived from the nature of the data-set: the number of images per patient is not constant across patients(for example patient A can have 10 images, and patient B can have 27 images). For this reason we designed our training loop based on stochastic gradient descent algorithm feeding one patient at a time and feeding all images for the patient so that all images are used for every patient. In order to reduce the variance on gradient estimates instead of doing a gradient descent update after processing every patient, we accumulate gradients and do an update after processing $k$ patients with $k$ being a parameter to our training loop thus simulating mini-batch gradient descent updates. On every epoch all patients are randomly sorted. 

A performance improvement for future work could be to statistically analyze the number of images per patient and select a fixed "window size" $w$(padding for patients with a number of images smaller than $w$ and truncating for patients with a number of images greater than $w$) or setting the window size $w$ to the maximum number of images per patient and padding for all the rest.

\paragraph{Patient Image Random Shuffle per Epoch}

On every epoch we perform a random shuffling of the images for every patient, this with the objective of not over-fitting to any particular sequence ordering. For the medical doctor something similar happens: he/she does not cares about any particular ordering, analyzes all the images and estimates diagnostic based on the set independently of the ordering. This means that for a given patient $i$ (called $patient_i$) the images belonging to $patient_i$  will be randomly ordered on every epoch before feeding them to the model. We hypothesize that this will create more robust hidden state feature vectors that should capture important/relevant information irrespective on the ordering the visual features are fed to the model. We leave for future word testing this hypothesis(and the impact on model performance) as well as  also experimenting with attention mechanisms and even models based purely on attention(for example: Transformers without positional encoding).

\subsection{Cerebral Hemorrhage Task using VGG19-LSTM}
For this task we combined a pre-trained VGG-19[5] convolutional neural network with a LSTM recurrent neural network and trained the combined model to predict cerebral hemorrhage on a sequence of images. This is the main task and the the task for which we report results and conclusions.

The architecture of the model is defined using the following modules:

\begin{itemize}
\item Pre-trained VGG-19[5] convolutional model having all layers frozen(non trainable) and last classification layer removed.
\item Added 2 fully connected(trainable) layers with 512 units each to the convolutional model, first fully connected layer having ReLU activation.
\item LSTM model with 1 hidden layer(600 units)
\item Linear layer connected to the output of the LSTM model
\item Output sigmoid layer with 1 unit for getting a probability of patient having cerebral hemorrhage given the image sequence($n$ images). $P(y=1|x_1,...,x_n)$
\end{itemize}
The model emits an output at every time-step $t$(every time-step is associated to an image in the sequence) interpreted as the probability of the patient having cerebral hemorrhage given the sequence of images processed up to time $t$ $P(y=1|x_1,...,x_t)$ but we use only the last time-step output(ignoring the rest) as prediction as well as in the weighted cross entropy(as implemented in [6]) loss function. On every time-step a feature vector is fed to the LSTM model obtained as the output vector of the convolutional model. The architecture can be illustrated as shown in \textbf{figure 2}  

\begin{figure}
    \centering
    \includegraphics[height=5cm]{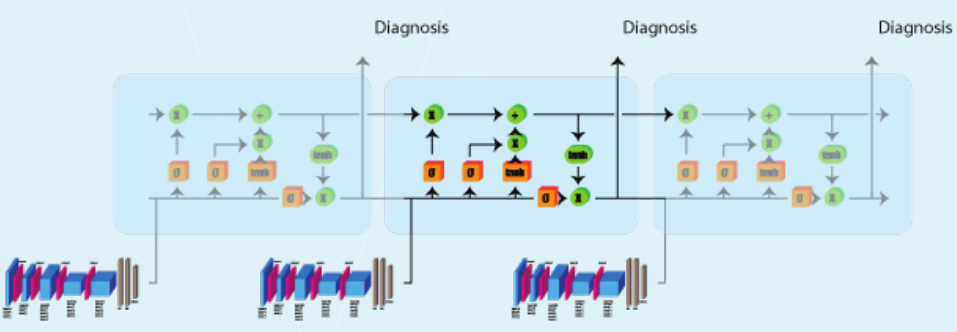}
    \caption{Cerebral Hemorrhage Task using VGG19-LSTM Model Architecture}
\end{figure}

\subsection{Multi-Diagnostic Task using Densenet121-LSTM}

This task served as a proof of concept on a limited data-set(the data-set was provided by batches to the team so it was not complete from the beginning) with the goal of over-fitting to it. Relevant details of the experimental setup are provided next:

\begin{itemize}
\item Pre-trained Densenet121[6] convolutional model having first 50 layers frozen(non trainable) and last classification layer removed.
\item Added 1 fully connected(trainable) layer with 512 units(no activation function)  to the convolutional model.
\item LSTM model with 1 hidden layer(512 units)
\item Output sigmoid layer with 5 units(1 per disease) for getting a probability of patient having cerebral hemorrhage given the image sequence($n$ images). $P(y_1=1|x_1,...,x_n),P(y_2=1|x_1,...,x_n),P(y_3=1|x_1,...,x_n),P(y_4=1|x_1,...,x_n),P(y_5=1|x_1,...,x_n)$
\end{itemize}
This model emits an output vector at every time-step t(every time-step is associated to an image in the sequence) where every element of the vector belongs to one of 5 diseases interpreted as the probability of that disease being present  being present for the patient in the sequence of images up to time $t$, we ignore all output vectors but the last one which is used as output of the model in the loss function and also for getting diagnostic predictions at inference time. Although this design provided good initial results(probably overfitted to the small data-set used) it showed to be computationally expensive having every experiment take several hours to run(on a single GPU) thus we decided to go for the simpler VGG19[5] model described previously.

\section{Results}
Many experiments were executed, here we list the best results including the architecture, hyper-parameters and training setup used .

\subsection{Cerebral Hemorrhage Task using VGG19-LSTM}

For this task the best experiment consists of 120 training epochs,the training cost curve (using weighted cross entropy[6]) decreases  reaching approximately 0.28 as seen in \textbf{Figure 3} 

\begin{figure}
    \centering
    \includegraphics[height=4cm]{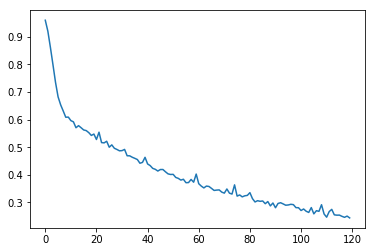}
    \caption{Training weighted cross-entropy loss for Hemorrhage VGG19-LSTM}
\end{figure}

The accuracy curves show the model is learning as expected reaching approximately 96\% accuracy in the training set and 85\% in the validation set. Given the skewness in the dataset we also evaluate other metrics including recall(sensitivity) ,precision (specificity) and f1-score. F1-score metrics reached approximately 0.95 in train set and 0.75 in validation set ,evaluation metric suggests over-fitting  (see \textbf{ figure 4}). We leave over-fitting reduction and generalization improvement for future work.

\begin{figure}
    \centering
    \includegraphics[height=10cm]{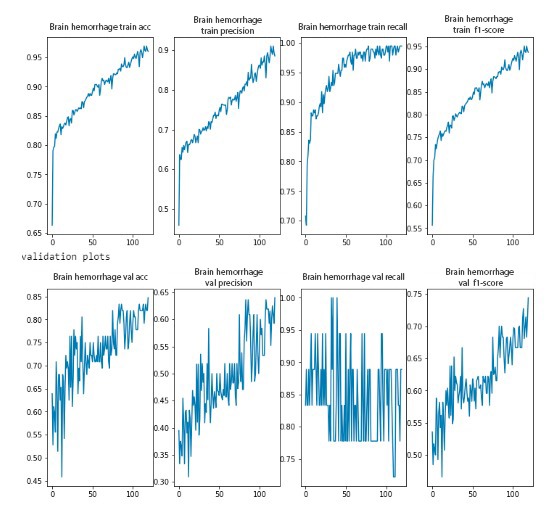}
    \caption{Evaluation metrics in train(row 1) and validation (row2) sets}
\end{figure}

\subsection{Multi-Diagnostic Task using Densenet121-LSTM}

For this task the best experiment consists of 300 training epochs, the training cost curve(for this task we used cross-entropy, instead of weighted cross entropy as described in the previous task) decreases reaching approximately 0.16 as seen in \textbf{figure 5}

\begin{figure}
    \centering
    \includegraphics[height=4cm]{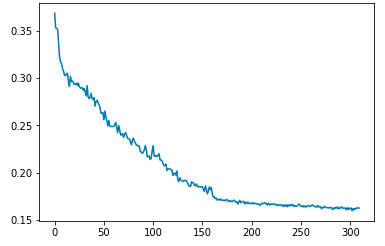}
    \caption{Training cross-entropy loss for Multi-Diagnostic Densenet121-LSTM}
\end{figure}

In terms of other evaluation metrics we measure a combined accuracy(averaged across the five diseases ) which increases as training goes on up to approximately 98\% as shown in \textbf{figure 6} 
\begin{figure}
    \centering
    \includegraphics[height=4cm]{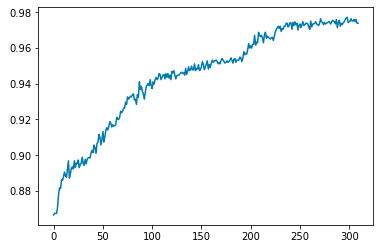}
    \caption{Training accuracy for Multi-Diagnostic Densenet121-LSTM}
\end{figure}

 Due to skewness present in the data-set other evaluation metrics were measured revealing that mass predictions were biased towards 0  due to the low prevalence of this disease, however for other cases like fracture and hemorrhage f1-score metric reveals promising results (close to 1) however this was measured just in the data-set so it is very probable these results are not of statistical significance due to overfitting(\textbf{figure 7}).
 
 \begin{figure}
    \centering
    \includegraphics[height=18cm]{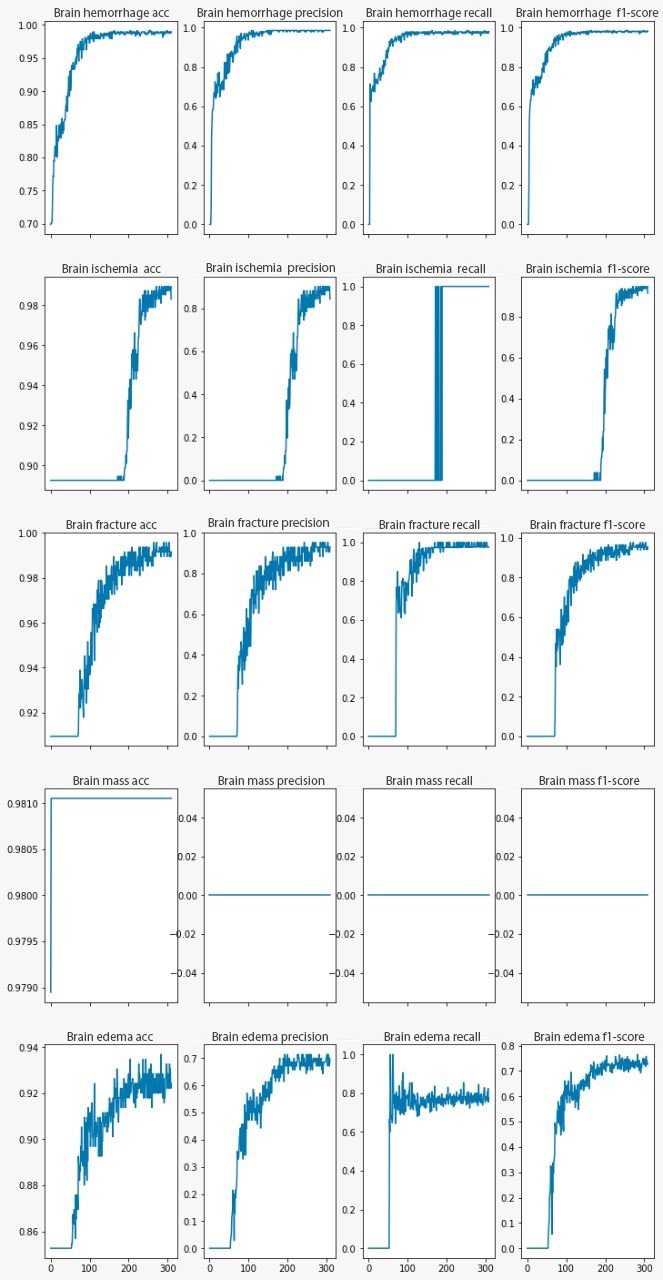}
    \caption{Training accuracy for Multi-Diagnostic Densenet121-LSTM}
\end{figure}
\section{Conclusions and Discussion}
\begin{itemize}
\item Experimental results supports the study hypothesis : \emph{If a vision model is combined with a memory sequence model , then it will be able to store relevant visual features from a sequence of images in order to emit a single diagnostic for the collection of images}. 

\item We tested this in a cerebral hemorrhage diagnostic task formulated as a multi-input to single binary classification problem where we have achieved approximately 95\% accuracy in the training set and 85\% in the validation set by combining a pre-trained convolutional model with a LSTM sequence model,which is enough for this study objectives. 

\item We leave for future work pushing the model performance to the highest   possible and more rigorous evaluation of generalization error in a held-out test set.

\item The custom data-set used is very small, and the computer resources limited to a personal computer with a single GPU, we  hypothesize better results can be obtained using a bigger and varied data-set and more computational resources to  train bigger state of the art models.
\end{itemize}

\section*{References}

[1] Sasank Chilamkurthy, et al. Development and Validation of Deep Learning Algorithms for Detection of Critical Findings in Head CT Scans. arXiv:1803.05854.

[2] LeCun, et al. Gradient Based Learning Applied to Document Recognition. Proceedings of the IEEE, vol. 86, no. 11, pp. 2278-2324, Nov. 1998, doi: 10.1109/5.726791.

[3] Sepp Hochreiter, et al. Long Short-term Memory. Neural computation. 9. 1735-80. 10.1162\/neco.1997.9.8.1735.

[4] Casamitjana, et al. 3D Convolutional Neural Networks for Brain Tumor Segmentation: A Comparison of Multi-resolution Architectures. 150-161. 10.1007/978-3-319-55524-9\_15.

[5] Karen Simonyan et al. Very Deep Convolutional Networks for Large-Scale Image Recognition. arXiv:1409.1556.

[6] Gao Huang  et al. Densely Connected Convolutional Networks. arXiv:1608.06993.

[7] Pranav Rajpurkar  et al. CheXNet: Radiologist-Level Pneumonia Detection on Chest X-Rays with Deep Learning. arXiv:1711.05225.

\end{document}